\documentclass[10pt,twocolumn,letterpaper]{article}

\usepackage{wacv}
\usepackage{times}
\usepackage{epsfig}
\usepackage{graphicx}
\usepackage{amsmath}
\usepackage{amssymb}

\usepackage{lineno,hyperref}
\modulolinenumbers[5]

\usepackage{subfig}
\usepackage{graphicx}
\usepackage{amsmath}
\usepackage{amssymb}
\usepackage{array}
\usepackage{booktabs}
\usepackage[linesnumbered,ruled]{algorithm2e}
\usepackage{algpseudocode}

\wacvfinalcopy 

\def\wacvPaperID{****} 

\ifwacvfinal\fi
\begin{document}

\title{Conditional Generative Refinement Adversarial Networks for Unbalanced Medical Image Semantic Segmentation}

\author{Mina Rezaei \\
Hasso-Plattner Institute\\
{\tt\small mina.rezaei@hpi.de}
\and
Haojin Yang \\
Hasso-Plattner Institute\\
{\tt\small haojin.yang@hpi.de}
\and
Christoph Meinel \\
Hasso-Plattner Institute\\
{\tt\small christoph.meinel@hpi.de}
}

\maketitle
\ifwacvfinal\thispagestyle{empty}\fi

\begin{abstract}
We propose a new generative adversarial architecture to mitigate imbalance data problem in medical image semantic segmentation where the majority of pixels are belong to healthy region and few belong to lesion or non-health region. A model trained with imbalanced data tends to bias toward healthy data which is not desired in clinical applications and predicted outputs by these networks have high precision and low sensitivity. We propose a new conditional generative refinement network with three components: a generative, a discriminative, and a refinement network to mitigate unbalanced data problem through ensemble learning. The generative network learns to segment at the pixel level by getting feedback from the discriminative network according to the true positive and true negative maps. On the other hand, the refinement network learns to predict the false positive and the false negative masks produced by the generative network that has significant value especially in medical application. The final semantic segmentation masks are then composed by the output of the three networks. The proposed architecture shows state-of-the-art results on LiTS-2017 for liver lesion segmentation, and two microscopic cell segmentation datasets MDA231, PhC-HeLa. We have achieved competitive results on BraTS-2017 for brain tumour segmentation.
\end{abstract}

\section{Introduction} \label{intro}
Medical imaging plays an important role in disease diagnosis, treatment planning, and clinical monitoring. One of the major challenges in medical image analysis is unbalanced data as normal or healthy data majority and lesion or non-healthy data are minor. A model learned from class imbalanced training data is biased towards the class with majority that is healthy. The predicted results of such networks have low sensitivity where sensitivity shows the ability of a test to correctly predict non-healthy classes. In medical applications the cost of miss-classification of the minority class could be more than the cost of miss-classification of the majority class. For example, the risk of not detecting tumour could be much higher than referring a healthy subject to doctors. 

The problem of class imbalanced have been recently addressed in diseases classification, tumour localization, and tumour segmentation and two types of approaches have been proposed in the literature: data-level approaches and algorithm-level approaches.

At the data-level, the objective is to balance the class distribution through re-sampling the data space~\cite{kohli2017medical} including SMOTE (Synthetic Minority Over-sampling Technique) of the positive class~\cite{douzas2018effective,mariani2018bagan} or under-sampling of the negative class~\cite{6914453}.
However, these approaches often lead to remove some important samples or add redundant samples to the training set.
Other techniques include iterative sampling~\cite{morales2012image} and incremental rectification of mini-batches for training deep neural network~\cite{dong2018imbalanced}.

Alternatively, algorithm-level based solutions address class imbalanced problem by modifying the learning algorithm to alleviate the bias towards majority class. Examples are accuracy loss~\cite{sudre2017generalised}, Dice coefficient loss~\cite{isensee2017brain,isensee2017automatic}, and asymmetric similarity loss~\cite{D11078} that modify distribution of training data with regards to miss-classification cost. These losses are able to cover only some aspects of the quality of the application. For example in case of segmentation different measures such as mean surface distance or Hausdorff surface distance need to be used.
Other approaches address balancing through ensemble learning by combining same or different classifiers to improve their generalization ability. The effect of combining redundant ensembles is studied by Sun et al.~\cite{sun2007cost} in term of bias and variance. The predicted results from the ensemble model improve in minority class due to a reduction in variance~\cite{sun2007cost}. In this work, we try to mitigate the negative impact of the class imbalance problem through ensemble learning from three networks of a generative, a discriminative, and a refinement.


Image segmentation is an important task in medical image computing which attempts to identify the exact boundaries of objects such as anatomical organs or abnormal regions. We apply our proposed method for automating medical image semantic segmentation. In our method, 3D bio-medical images are represented as a sequence of 2D slices (such as z-stacks). A long short-term memory (LSTM) is an effective unit for processing sequential data in order to exploit a long term temporal correlation. Bidirectional LSTMs~\cite{graves2005framewise} are an extension of classical LSTMs which are able to improve model performance on sequence processing. Bidirectional LSTMs have an advantage to access information in next slice as well as previous slice. This provides additional context and eliminate ambiguity from the network and result in faster learning~\cite{graves2005framewise}.
We utilize bidirectional LSTM units to enhance temporal consistency and get inter and intra-slice representation of features inside of the generative network, the discriminative network, and the refinement network.

\begin{figure}[!t]
\includegraphics[width=0.5\textwidth]{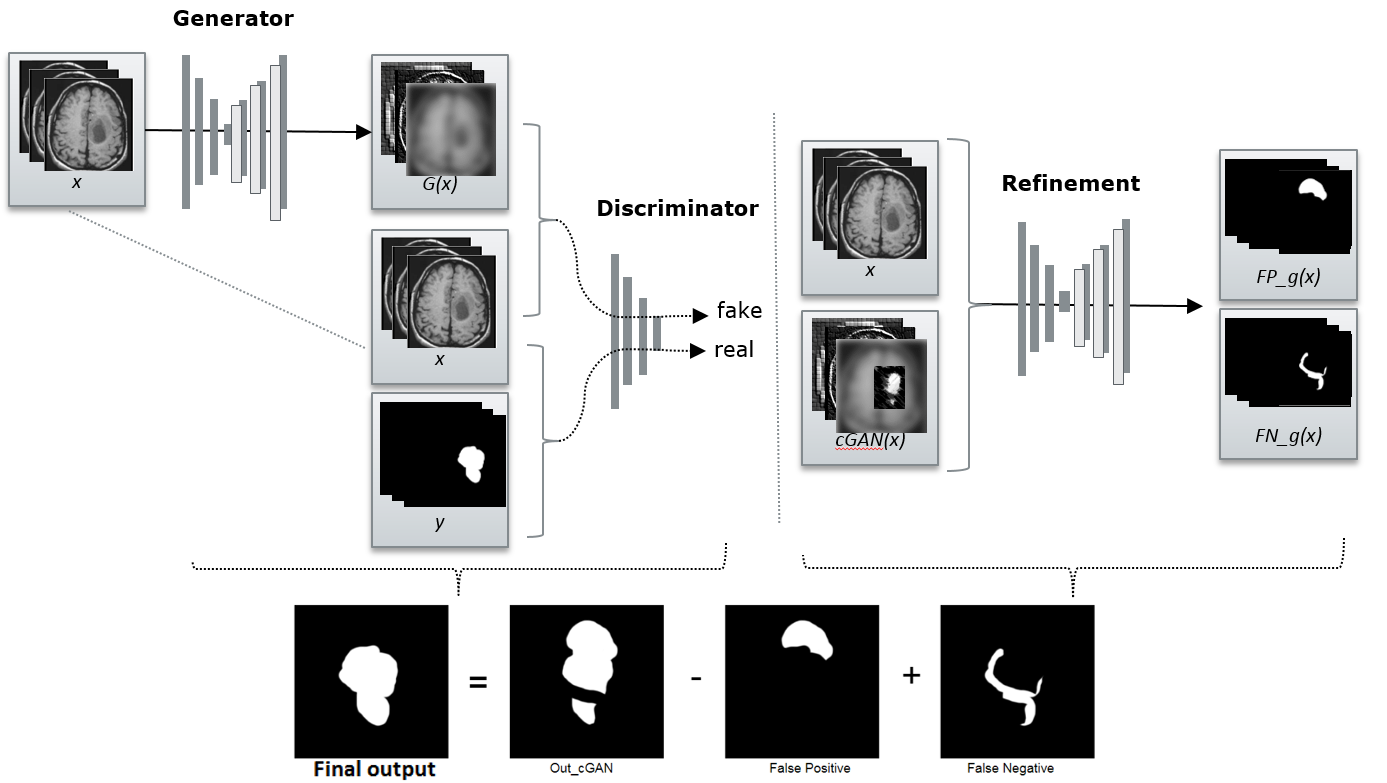}
\centering
\caption{The proposed method for medical image semantic segmentation consists of a generator network, a discriminator network, and a refinement network. The generator tries to segment image in pixel level, while discriminator classifies the synthesized output is real or fake. The final semantic segmentation masks are computed through eliminating the false positives and adding the false negatives predicted masks by the refinement network.}
\label{fig_model}
\end{figure}

Fig.~(\ref{fig_model}) shows our proposed method in two stages of a cGAN and a refinement network. The training procedure for the generator and the discriminator is similar to a two-player mini-max game, where a generator network and a discriminator network are trained in an alternating fashion to respectively minimize and maximize an objective function.
The generator takes 2D sequences of multi-modal medical images as condition and tries to generate corresponding segmentation labels. The discriminator determines the generator output is real or fake.
The refinement network learns false negative and false positive of the predicted masks produced by cGAN. The final semantic segmentation masks are computed by predicted masks from the cGAN and the refinement network.

We conducted experiments on different medical imaging benchmarks, which demonstrate the generalization ability of our approach for segmentation of body organ and tumorous region. 
The contributions of this work can be summarized as follows:
\begin{itemize}

\item We propose a conditional refinement GAN to mitigate imbalanced data issue for medical image semantic segmentation through ensemble learning~(Section \ref{methodology}).

\item We design the refinement network to tackle with miss-classification cost that has significant value especially in medical application~(Section \ref{methodology}).

\item We study the effect of different architectural choices and normalization techniques~(Section \ref{methodology} and \ref{experiment}).


\end{itemize}

\section{Methodology} \label{methodology}
In this section, we present the conditional refinement GAN for medical image semantic segmentation. To tackle with miss-classification cost and mitigate imbalanced medical imaging data, we proposed an ensemble network consists of a cGAN and a refinement.

\subsection{Conditional Refinement GAN} \label{CRGAN}
In a conventional generative adversarial network, generative model $G$ tries to learn a mapping from random noise vector $z$ to output image $y$; $G: z \rightarrow y$ .
Meanwhile, a discriminative model $D$ estimates the probability of a sample coming from the training data $x_{real}$ rather than the generator $x_{fake}$.
The GAN objective function is a two-player mini-max game like Eq.(\ref{eq_GAN}).

\begin{equation} \label{eq_GAN}
\begin{split}
 \underset{G} min \,  \underset{D } max \, V(D, G)
   = E_{y} [log D(y)] + \\
     E_{x,z} [log (1-D(G(x,z)))] 
\end{split}
\end{equation}

Unlike previous conditional GANs~\cite{MirzaO14,Phillipimagetoimage2017,XueXZLH17,Moeskops2017,KohlBSYHHRM17}; in our proposed method, a generative model learns mapping from a given sequence of 2D multimodal MR images $x_i$ to a sequence semantic segmentation $y_{seg}$; $G : \{x_i,z\} \rightarrow \{y_{seg}\}$ (where $i$ refers to 2D slice index between 1 and 155 from a total 155 slices acquired from each patient).
We utilize bidirectional LSTM to pass the temporal consistency between 2D slices. Our network is able to learn representations from previous and future slices which results context aware and eliminate ambiguity. 
The training procedure for the segmentation task is similar to two-player mini-max game as shown in Eq.(\ref{eq_svGAN}).
While the generator segmented pixels label, the discriminator takes the ground truth, and the generator's output to classify the output is real or fake.

\begin{equation} \label{eq_svGAN}
\begin{split}
\mathcal{L}_{adv} \leftarrow \underset{G} min \,  \underset{D } max \, V(D, G)
   = E_{x,y_{seg}} [log D(x,y_{seg})] + \\
     E_{x,z} [log (1-D(x, G(x,z)))]
\end{split}
\end{equation}

The generative loss~Eq.(\ref{G}) is mixed with $\ell1$ term to minimize the absolute difference between the predicted value and the existing largest value. Previous studies~\cite{Phillipimagetoimage2017,XueXZLH17} on cGANs have shown the success of mixing the cGANs objective with $\ell1$ distance.
The $\ell1$ objective function takes into account CNN feature differences between the predicted segmentation and the ground truth segmentation and resulting in fewer noises and smoother boundaries.

\begin{equation} \label{G}
 \mathcal{L}_{L1}(G) = E_{x,z} \parallel y_{seg} - G(x,z) \parallel  
\end{equation}

The adversarial loss for semantic segmentation task calculate by Eq.(\ref{adv})

\begin{equation} \label{adv}
\mathcal{L}_{seg} (D, G) = \mathcal{L}_{adv} (D, G) + \mathcal{L}_{L1}(G) 
\end{equation}

\begin{figure}[!t]
\includegraphics[width=0.45\textwidth]{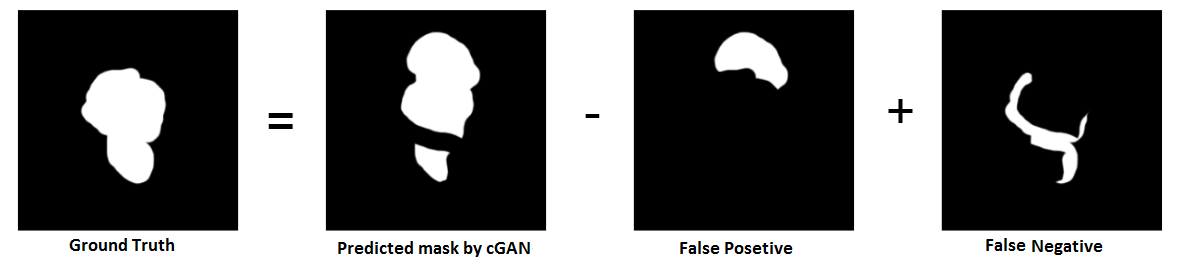}
\centering
\caption{Visual results from our model where the cGAN over segment through learning true positives and true negatives and the refinement learns false positives and false negatives mask.}
\label{fig_mask}
\end{figure}

As mentioned in Section~\ref{intro}, in order to tackle with miss-classification cost, the predicted output by the generator and discriminator are passed to refinement network. The refinement network is trained to learn the false prediction of cGAN in details of false negatives~(Eq.~\ref{eq_FN}) and false positives (Eq.~\ref{eq_FP}).  
The false negative error represents the number of pixels that were incorrectly labeled as background or wrong class (Fig.~(\ref{fig_mask}) third column). Similarly, the false positive indicates the number of pixels that were incorrectly labeled as part of the region of interest (Fig.~(\ref{fig_mask}) last column).

\begin{equation} \label{eq_FN}
\mathcal{L}_{fn} =  clip {((y - {\mathcal{L}_{seg} )}, 0, 1)} 
\end{equation}

\begin{equation} \label{eq_FP}
\mathcal{L}_{fp} =  clip {(({\mathcal{L}_{seg}- y)} , 0, 1)} 
\end{equation}

where in both equations (\ref{eq_FN} and \ref{eq_FP}) $y$, $\mathcal{L}_{seg}$ respectively refers to the ground truth labels and predicted labels by adversarial loss.

Our final objective function $\mathcal{L}_{CR-GAN}$ for semantic segmentation relies on adding false negatives and subtracting false positives from outputs of adversarial network. 
\begin{equation} \label{final_loss}
\mathcal{L}_{CR-GAN} = {\mathcal{L}_{seg} - \mathcal{L}_{fp} + \mathcal{L}_{fn}  } 
\end{equation}


\subsection{Network Architectures}  \label{arch}

As shown in Fig.~(\ref{fig_model}), our proposed method consists of a generator network, and a discriminator network, in the left side followed by a refinement network in the right side of the figure.
We investigate two different architectures of conditional GAN~(Section \ref{method_sec1a}) and recurrent conditional GAN~(Section \ref{method_sec1a}) for adversarial training of $G$ and $D$.

\subsubsection{Conditional Generative Adversarial Network} \label{method_sec1a}
In our cGAN architecture, the generator is a fully convolutional encoder-decoder network that generates a label for each pixel.
Similar to UNet~\cite{ronneberger2015u}, we added the skip connections between each layer $i$ and layer $n-i$, where $n$ is the total number of layers.
Each skip connection simply concatenates all channels at layer $i$ with those at layer $n-i$.
We use the convolutional layer with kernel size 5 $\times$ 5 and stride 2 in encoder part for down-sampling, 
and in decoder section  perform up-sampling by image re-size layer with a factor of 2 and convolutional layer with kernel size 3 $\times$ 3 stride 1.
In our architecture, in last layer, the high resolution features from multi-modal, multi-site images are concatenated with up-sampled versions of global low-resolution features which helps the network learn both local and global representation of features.

The discriminator is a fully convolutional networks and has same architecture as decoder part of generator network. The hierarchical features from convolutional layers passed to softmax loss for classifying whether a segmented pixel's label belongs to right class.

\subsubsection{Recurrent Generative Adversarial Networks} \label{method_sec1b}
Similar to our cGAN, in the recurrent cGAN both generator and discriminator substitutes with bidirectional LSTM units~\cite{graves2005framewise}.
The recurrent conditional GAN has an advantage of getting temporal consistency between previous and next slice.
Using bidirectional LSTM units inside of $G$ and $D$ makes networks context aware, which is an important point in temporal data analysis. 

\subsubsection{Refinement Network} \label{refarch}

We design the refinement network on top of adversarial network to deal with unbalanced data issue and improve classification. The refinement network is UNet architecture with bidirectional LSTM in circumvent of bottleneck which takes a 2D sequence outputs from cGAN (or recurrent cGAN), with a 2D sequence of medical images, and the outputs are a 2D sequence masks of false positives and false negatives.

The final semantic segmentation extracted by adding false negatives and subtracting false positives predicted by refinement from outputs of cGAN network.

All proposed architectures in this paper apply a patient-wise mini-batch normalization technique described in the subsection~(\ref{batch_norm}).

\subsection{Patient-wise Batch Normalization}  \label{batch_norm}
Several popular techniques are developed for normalization, such as batch normalization~\cite{IoffeS15}, and max norm constraints~\cite{srebro2005rank}, with the core idea of shifting the inputs to a zero mean and unit variance. The inputs are normalized before applying non-linearity to prevent the inputs from saturating extreme non-linearity. As described by Ioff~\textit{et al}.~\cite{IoffeS15}, batch normalization improve the overall optimization and gradient issues. In many cases, initial weights have a large deviance from true weights, delaying the convergence during training. Batch norm reduces the influence of weight deviance by normalizing the gradients this speed up the training.

Recently, stratified batch sampling is shown successful results in personalized medicine~\cite{kim2013stratified} and statistic~\cite{keramat1998study} when sub-populations within an overall population are vary. Stratified sampling can reduce variance~\cite{zhao2014accelerating} through sampling each sub population (stratum) independently where the strata are constructed within homogeneous and among heterogeneous.

Similar to the concept of stratified sampling, we initially normalized the inputs where the mean and variance are computed on a specific patient from the same acquisition plane (Sagittal, Coronal, and Axial) and from all available image modalities (e.g., T1, T1-contrast, T2, Flair in the BraTS benchmark).
In this regard, the deviances get increasingly large, and the back-propagation step needs to account for these large deviances which this restrict us from using a small learning rate to prevent gradient explosion. For example, the mini-batch with 128 images includes the same patient images and four available modalities from the same acquisition plane. Algorithm.~\ref{minibatch_alg}, shows how to compute normalization at each mini-batch by proposed patient-wise batch-norm technique.

\begin{algorithm} 
    \SetKwInOut{Input}{Input}
    \SetKwInOut{Output}{Output}

    \Input{{Values of $x$ over a mini-batch: $\beta={x_1, x_2, ...,x_{155}}$} \\
           {Parameters to be learned:${\gamma,\beta}$}
           }
    \Output{$y_i= BN_{\gamma,\beta} (x_i)$}
    
    \For{$Patient:{P_1, P_2, ..., P_n} $}
    {
    \For{$AcquisitionPlane:{x_i, y_i, z_i}$}
    {
    \For{$Image \quad Modalities:{T1, T2, T1c, Flair}$}
    {
    $\mu _{\beta}   \leftarrow   \frac{1}{m} {\sum\limits_{i=1}^n {x_{i}}} $ \\
    $\sigma^2_{\beta}  \leftarrow   \frac{1}{m} {\sum\limits_{i=1}^n {(x_{i} - \mu_\beta})^2} $ \\
    $\hat{x_i}  \leftarrow   \frac{x_i - \mu_x}{\sqrt[2]{\sigma^2_{x} + \varepsilon  }} $ \\
    $y_{i}  \leftarrow \gamma \hat{x_i} + \beta  = BN_{\gamma,\beta} {(x_i)}$
    }
    }
    }
 
\caption{Patient-wise mini-batch normalization. ($i$ and $n$ respectively refer to a number of 2D slices and number of patient e.g. 0 $<$ $i$ $\leq$ 155, $n$=230 in BraTS)}
\label{minibatch_alg}
\end{algorithm}

\section{Experiments}  \label{experiment}
To evaluate the performance of our network on imbalanced data segmentation and compared it with state-of-the-art methods, we trained recent popular annotated medical imaging benchmarks as described in~ Section (\ref{datasets}).

\subsection{Dataset and Pre-processing} \label{datasets}

The first experiment is carried out on real patient data obtained from BraTS2017 challenge~\cite{Menze2014,Bakasnature2017,Bakastcg2017,Bakaslgg2017}.
The BraTS2017 released data in three subsets train, validation, and test comprising 289, 47, and 147 MR images respectively in four multisite modalities of T1, T2, T1ce, and Flair which the annotated file provided only for the training set. The challenge is semantic segmentation of complex and heterogeneously located of tumour(s) on highly imbalanced data. Pre-processing is an important step to bring all subjects in similar distributions, we applied z-score normalization on four modalities with computing the mean and stdev of the brain intensities. We also applied bias field correction introduced by Ny{\'u}l et al.~\cite{nyul2000new}.

In second experiment, We applied the LiTS2017 benchmark which contains 130 computer tomography (CT) training data, and 70 test set. The examined patients were suffering from different liver cancers. The challenging part is semantic segmentation of unbalance labels with a large (liver) and small (lesion) target. Here, pre-processing is carried out in a slice-wise fashion. We applied Hounsfield unit (HU) values, which were windowed in the range of [−100, 400] to exclude irrelevant organs and objects. Furthermore, we applied histogram equalization to increase the contrast for better differentiation of abnormal liver tissue.

In third experiment, we test the performance of our proposed method on small size microscopic light dataset from human breast carcinoma cells. 
Additionally, we provided data augmentation such as randomly cropped, re-sizing, scaling, rotation between -10 and 10 degree, and Gaussian noise applied on training and testing time for three datasets.

\subsection{Implementation}

\textbf{3.2.1. Configuration:} Our proposed method is implemented based on a Keras library~\cite{chollet2015keras} with backend Tensorflow~\cite{abadi2016tensorflow} and our code is publicly available~\footnote{https://github.com/anonymous}. We did not use any pre-trained model in our experiments and started training from scratch. All training and experiments are conducted on a workstation equipped with couple NVIDIA TITAN X GPUs. The learning rate is initially set to 0.001. The RMSprop optimizer is used in the recurrent generator, discriminator, and refinement, it dividing the learning rate by an exponentially decaying average of squared gradients. We used Adadelta as an optimizer for cGAN network that continues learning even when many updates have been done.

The cGAN, recurrent cGAN, and refinement model are trained separately for up to 100 epochs. In this work, the recurrent architecture selected for both discriminator and generator is a bidirectional LSTM proposed by Graves et al.~\cite{graves2005framewise}. We used all 2D sequences from axial, coronal, and sagittal planes from the both training and testing phases.

\textbf{3.2.2. Network Architecture:} In this work, a generator network is a modified UNet architecture with bidirectional LSTMs unit. The UNet architecture allows low-level features to shortcut across the network. The bidirectional LSTM provides inter as intra slice feature representation which is very important in sequential medical image analysis. The advantage of bidirectional LSTM appear when we connected features from $n-1-i$ and $i$ (where n refers to total of layers).

Our discriminator is fully convolutional Markovian PatchGAN classifier~\cite{Phillipimagetoimage2017} which only penalizes structure at the scale of image patches.
Unlike, the PathGAN discriminator introduced by Isola et al.~\cite{Phillipimagetoimage2017} which classified each N × N patch for real or fake, we have achieved better results for task of semantic segmentation in pixel level where we consider N=1. Moreover, since we have a sequential data, the bidirectional LSTM added after last CNN layer in discriminator network. We used categorical cross entropy~\cite{nasr2002cross} as an adversarial loss with combination of $\ell1$ loss in generator network.

Regarding the highly imbalance datasets, minority pixels with lesion label are not trained as well as majority pixels with non-lesion label. Therefore, we designed refinement network to tackle this issue. The refinement network has same architecture as our recurrent generator. The refinement network takes the predicted output from cGAN and medical images. The refinement network outputs two binary masks: false positive and false negative. 

\subsection{Evaluation Results and Discussion} \label{evaluation}
The quantitative evaluation and comparison was based on the online judgment system provided by BraTS2017 challenge~\footnote{http://braintumorsegmentation.org/}.
We also evaluated the performance of our approach on CT images for semantic segmentation of liver and lesion using the quality metrics introduced in the LiTS2017 from grand challenges~\cite{heimann2009comparison}. 

\textbf{3.3.1 Heterogeneous Brain Tumor Segmentation:}

The segmentation of the brain tumour from medical images is highly interesting in surgical planning and treatment monitoring.
The goal of segmentation as described by organizer~\cite{Menze2014,Bakasnature2017,Bakastcg2017,Bakaslgg2017} is to delineate different tumour structures such as active tumorous core~(TC), enhanced tumorous~(ET), and edema or whole tumorous~(WT) region.

\begin{figure*}[!htbp]
\includegraphics[width=0.89\textwidth]{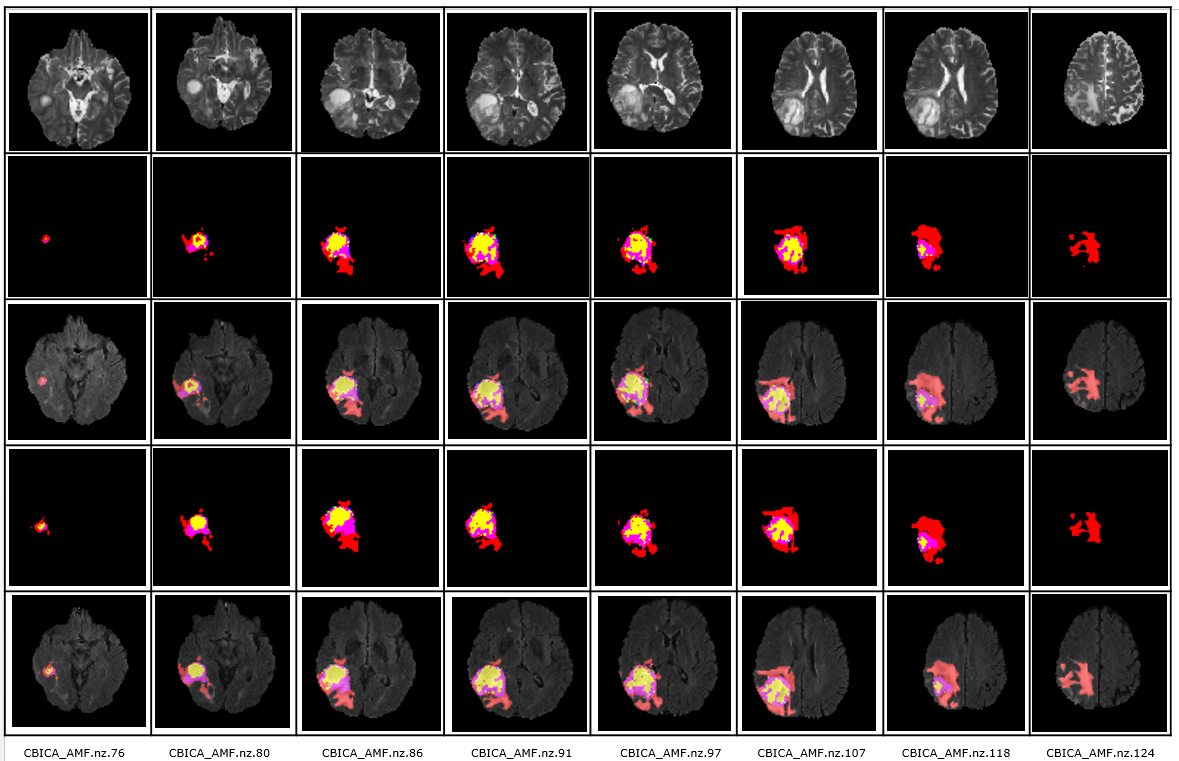}
\centering
\caption{Visual results from our model on axial views of CBICA-AMF.nz.76-124 from the validation set. The first row shows Flair modality, while the second and fourth row show the output results respectively from cGAN and refinement architecture. The third row shows the semantic segmentation masks from cGAN overlaid Flair modalities where the fifth row shows outputs after refinement network. The red color codes the whole tumour (WT) region, while pink and yellow represent the enhanced tumour (ET) and the tumorous core (TC) respectively.}
\label{fig_brain_ex1}
\end{figure*}

Fig.~(\ref{fig_brain_ex1}) shows qualitative results of the cGAN network, and refinement network in detail. Based on Fig.~(\ref{fig_brain_ex1}), the result shows good relation to the ground truth for the segmentation after refinement network. The final output is refined through eliminating false negative pixels, and adding the false positive pixels.

The Dice score, Hausdorff distance, sensitivity, and specificity are introduced by BraTS2017 as evaluation criteria for segmentation task.
Tables~(\ref{tableDiceHauseresult}, \ref{tableSenSperesult}) present the brain segmentation results from proposed architecture and compare them with other related methods based on the pre-proceeding report~\cite{BRATS2017preproceeding}.

\begin{table*}[!htbp]
\centering
\caption{Comparison of the achieved accuracy for semantic segmentation of different classes of tumour in terms of Dice and Hausdorff distance on validation data~\cite{Menze2014,Bakasnature2017,Bakastcg2017,Bakaslgg2017} reported by the BraTS2017 organizer. The terms WT, ET, and TC are abbreviations of whole tumor region, enhanced tumor region, and core of tumor respectively.}
\begin{tabular}{*7c} 
\toprule
Label    &  Dice-WT   & Dice-ET    & Dice-TC    & Hdf-WT  &  Hdf-ET  & Hdf-TC \\
\midrule
RNN-cGAN+Refinement & 0.86 & 0.64 & 0.73 & 7.22 & 8.30 & 11.04 \\
cGAN & 0.74 & 0.53 & 0.61 & 12.6 & 16.41 & 31.0 \\
Recurrent-cGAN & 0.79 & 0.60 & 0.68 & 11.73 & 14.54 & 25.83 \\
Residual-Encoder~\cite{Pawar17} & 0.82 & 0.62 & 0.57 & - & - & - \\
FCN~\cite{Alex1717} & 0.83 & 0.69  & 0.69 &  11.06 & 11.49 & 12.53 \\
3D-Unet~\cite{Amorim17} &  0.81 & 0.76  & 0.72 &  13.65 & 22.36  & 13.88 \\
Masked-Vnet~\cite{cata17} & 0.86 & 0.71 & 0.63 & 5.43 & 8.34  & 11.17 \\
3D-Seg-Net~\cite{Dong17} &  0.79 & 0.60  & 0.64 &  23.33 & 21.09  & 26.01 \\
Nifty-Net~\cite{EatonRosen17} & 0.83 & 0.71  & 0.68 &  27.49 & 17.35  & 31.34 \\
3D-CNN~\cite{German17} &  0.82 & 0.46 & 0.56 & 9.56 & 13.8 & 14.7 \\
biomedia~\cite{pawlowski2018ensembles} &  0.90 & 0.73 & 0.79 & 4.2 & 4.5 & 6.5 \\
UCL-TIG~\cite{wang2017automatic} &  0.90 & 0.78 & 0.83 & 3.8 & 3.2 & 6.4 \\
MIC-DKFZ~\cite{isensee2017brain} &  0.89 & 0.73 & 0.79 & 6.9 & 4.5 & 9.4 \\
\bottomrule
\end{tabular}
\label{tableDiceHauseresult}
\end{table*}

\begin{table*}[!htbp]
\centering
\caption{Comparison and the achieved accuracy for semantic segmentation in terms of false negative rate or FNR= $1- \frac {TruePositive} {TruePositive+FalseNegative}$ and false positive rate or FPR=$1 - \frac {TrueNegative} {TrueNegative+FalsePositive}$ on validation data. The terms of WT, ET, and TC are abbreviations of whole tumor region, enhanced tumor region, and core of tumor respectively.}
\begin{tabular}{*7c} 
\toprule
Label    &  FNR-WT   & FNR-ET   & FNR-TC &  FPR-WT   & FPR-ET   & FPR-TC   \\
\midrule
RNN-cGAN+Refinement  &  0.11 & 0.16  & 0.29  & 0.02 & 0.02 & 0.02 \\
cGAN  &  0.22 & 0.34 & 0.32  &  0.02  &  0.04  & 0.03   \\
Recurrent-cGAN  &  0.19 & 0.32 & 0.30  &  0.02  &  0.03  & 0.02   \\
biomedia~\cite{pawlowski2018ensembles} &  0.11 & 0.22 & 0.24  & - & -  & - \\
UCL-TIG~\cite{wang2017automatic}& 0.09 & 0.23 & 0.18 & - & - & - \\
MIC-DKFZ~\cite{isensee2017brain} & 0.11 & 0.21 & 0.22 & - & - & - \\
\bottomrule
\end{tabular}
\label{tableSenSperesult}
\end{table*}

From Table (\ref{tableDiceHauseresult}), the cGAN network (in second line) with one generator and discriminator achieved 12\% less accuracy for whole tumour region segmentation compared to the segmentation results after the refinement network.
In the first stage, the generator is trained by true positive and true negative masks.
Meanwhile, the discriminator network tests how true is the predicted mask created by the generator.
On the top of cGAN, the refinement learns the false negative and false positive masks.
Table~(\ref{tableSenSperesult}) presents discovery of false negative rate (1-recall) and false positive rate (1-specificity) in detail of network architecture.
The final masks computed from the cGAN (or recurrent-cGAN) network with eliminating false negative and adding false positive predicted by refinement network.

Regarding results of false discovery rate presented in Table (\ref{tableSenSperesult}), we have achieved good results as second and third ranked teams in BraTS2017 competition when the segmented masks computed by recurrent conditional GAN and refinement network. 
Regarding quantitative results by Tables (\ref{tableDiceHauseresult} and \ref{tableSenSperesult}), the networks substituted by LSTM unit predicted more accurate results.


In test time, every group had 48 hours from receiving the test subjects to process them and submit their segmentation results to the online evaluation system.
The average value of the Dice coefficient is 0.85 in test time, which the results from Table~(\ref{table-validation-test-result}) obtained and evaluated by challenge organizer. Since the results of the challenge in testing are not publicly available, we are not able to compare the performance of the different approaches in the test time.

\begin{table}[!htbp]
\centering
\caption{The achieved accuracy for brain tumour semantic segmentation by proposed conditional refinement GAN in terms of Dice, sensitivity, specificity, and Hausdorff distance reported by the BraTS-2017 organizer.}
\begin{tabular}{*7c} 
\toprule
Evaluation  &  \multicolumn{3}{c}{Validation} & \multicolumn{3}{c}{Test}\\
\midrule
{}     &  WT   & ET    & TC    & WT  &  ET  & TC \\
Dice   &  0.86 & 0.64  & 0.73  &  0.85 & 0.61  & 0.72 \\
Sens   &  0.89 & 0.84  & 0.71  &  -    &  -    &  -   \\
Spec   &  0.98 & 0.98  & 0.97  &  -    &  -    &  -   \\
Hdfd   &  7.22 & 8.30  & 11.04 &  8.73 & 59.2  & 25.9 \\
\bottomrule
\end{tabular}
\label{table-validation-test-result}
\end{table}

It is important to mention that our method takes only 58 seconds to segment one MR brain image consisting 155 slices at testing time.

\textbf{3.3.2 Simultaneous Liver and Lesion(s) Segmentation:}

Liver cancer is one of the most common types of cancers around the world~\cite{cancers6010226} and CT images are widely used for diagnosis of hepatic diseases. The proposed method was trained on the public clinical CT dataset from LiTS2017 competition.

\begin{figure*}[!t]
  \centering
  \subfloat[]{\includegraphics[width=0.50\textwidth]{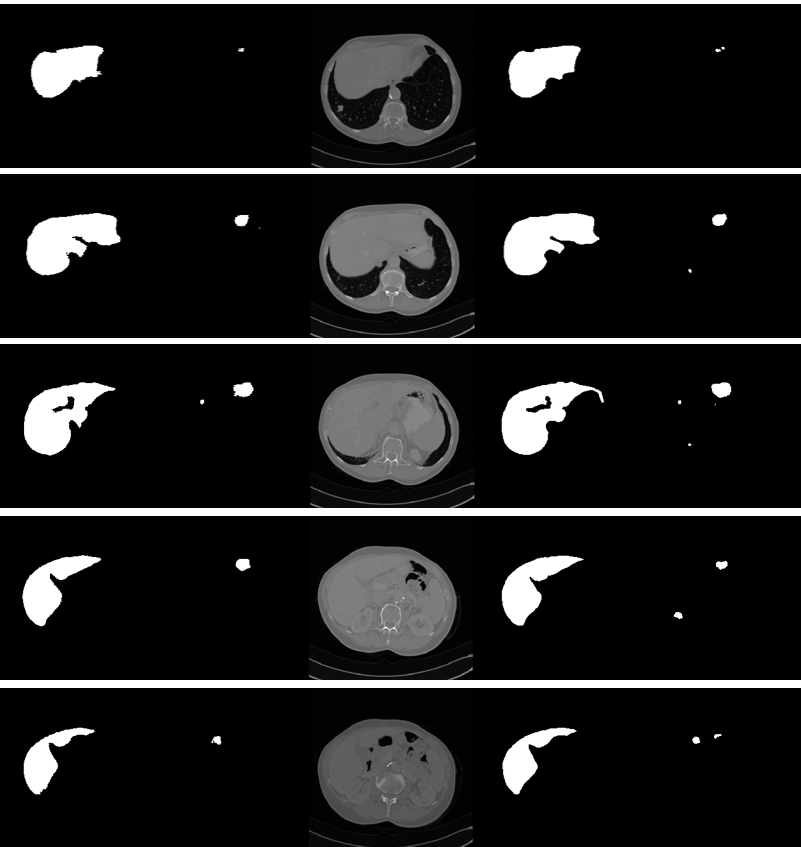}\label{fig_1a}}
  \hfill
  \subfloat[]{\includegraphics[width=0.47\textwidth]{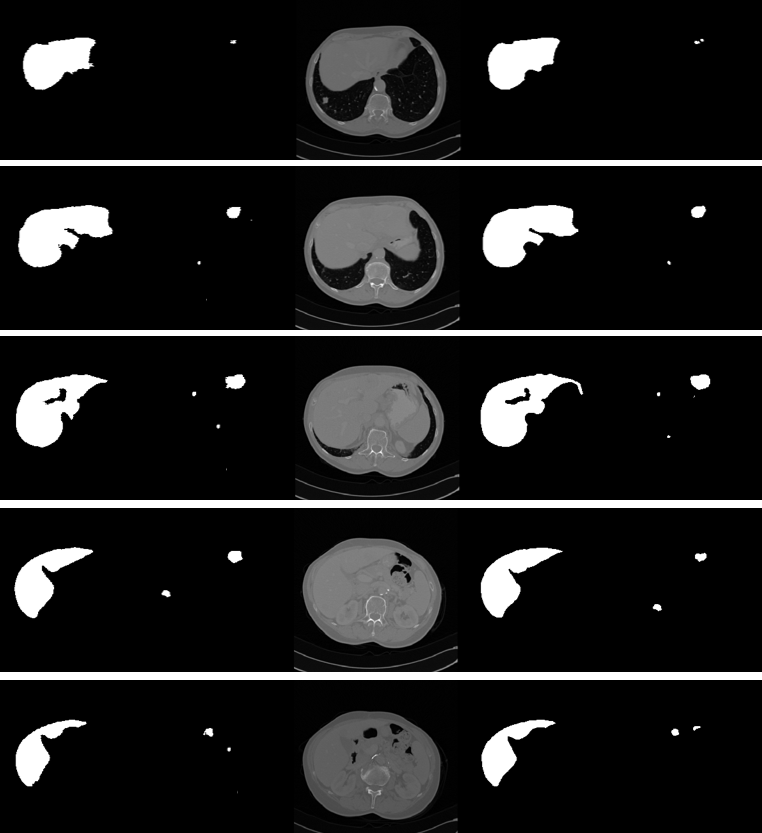}\label{fig_2b}}
  \caption{Segmentation results obtained by cGAN (a) compared to the refinement output (b). In each sub figure, the first two left columns show the ground truth manual segmentation of the liver and lesion(s). The two last right columns from (a,b) show the predicted liver and lesion(s) at the first and second stages.}
\label{fig_livres}
\end{figure*}

Fig.~(\ref{fig_livres}) shows segmentation output in detail of conditional GAN in the left followed by refinement output in the right side of figure.

In this competition the primary metric is the Dice score. A volume overlap error (VOE), relative volume difference (RVD), average symmetric surface distance (ASSD), and maximum symmetric surface distance (MSSD) are considered for the evaluation of predicted region of liver and lesion(s). Tables~(\ref{dice_asd_liver} and \ref{ourmertricliver}) describe the quantitative results and comparisons with top ranked methods from LiTS leader-board~\footnote{https://competitions.codalab.org/competitions/}.

\begin{table}[!htbp]
\centering
\caption{The achieved accuracy for simultaneous liver and lesions segmentation in terms of Dice score and average surface distance on the test data where the 1 is index of a liver and 2 for a lesions.}
\begin{tabular}{*5c} 
\toprule
Approaches & Dice1  & Dice2 & ASD1 & ASD2 \\
\midrule
cGAN+Refinement   &  0.94 & 0.83  & 1.4 & 1.6 \\
cGAN   &  0.85 & 0.81  & 1.8  &  2.1  \\
UNet   &  0.72 & 0.70  & 19.04  &  19.04    \\
ResNet+Fusion~\cite{BiKKF17} & 0.95 & 0.50  & 0.84 &  13.33  \\
SuperAI & 0.96  & 0.81  & -  &  1.1   \\
H-Dense+ UNet~\cite{Han17a} &  0.96 & 0.82  & 1.45  & 1.1    \\
coupleFCN~\cite{VorontsovCTPK17} &  0.78 & 0.77  & -  &  -    \\
\bottomrule
\end{tabular}
\label{dice_asd_liver}
\end{table}

\begin{table}[!htbp]
\centering
\caption{The top two rows show achieved accuracy for the task of simultaneous liver and lesions segmentation in terms of Dice score and average surface distance on the test data.}
\begin{tabular}{*5c} 
\toprule
Architecture & VOE  & RVD & ASD & MSD \\
\midrule
cGAN+Refinement   &  14 & -6  & 6.4 & 40.1 \\
cGAN   &  21 & -1  & 10.8  &  87.1  \\
ResNet+Fusion~\cite{BiKKF17} & 16 & -6  & 5.3 &  48.3 \\
SuperAI & 36 & 4.27  & 1.1 &  6.2 \\
H-Dense+ UNet~\cite{Han17a} &  39 & 7.8  & 1.1 &  7.0 \\
coupleFCN~\cite{VorontsovCTPK17} & 35 & 12  & 1.0 &  7.0 \\
\bottomrule
\end{tabular}
\label{ourmertricliver}
\end{table}

To have better understanding about the performance gains, we analyze the achieved accuracy on imbalanced liver tumor segmentation dataset where we can see unbalancing labels between large body organ and very small lesions.
Based on the leader-board, most top ranked models used cascade networks to segment simultaneously~\cite{Han17a} or separately~\cite{BiKKF17, VorontsovCTPK17} liver as well as lesion.
The cascade networks provide good solution against imbalanced labeling.

Table~(\ref{dice_asd_liver}) describes our obtained result for liver segmentation and lesions in terms of the Dice score 0.94 and 0.83 respectively.
Based on Table~(\ref{dice_asd_liver}) and with comparison of the first two rows, we can see the effect of refinement network on final results which has increased up to 9\% for liver segmentation and similarly up to 2\% for the lesions segmentation.

In the LiTS dataset, lesions with an approximate diameter equal to or larger than 10 mm was defined as a large one, while a small lesion has a diameter of less than 10 mm.
Our method achieved an average Dice of 0.90 and ASD of 1.6 in lesion segmentation which obviously, can distinguish small and large lesions. 

In addition, our algorithms are very fast, and it takes only 100 seconds for the simultaneous segmentation of liver and lesion from CT images with 280 slices, each sized 512 x 512.
The complex and heterogeneous structures of the predicted liver and all lesions from local test set are depicted in Fig.~(\ref{fig_livres}).

\textbf{3.3.3 Microscopic Cell Segmentation:}

Microscopy cell images are key component of the biological research process and automatic cell segmentation is helpful application for clinical routine. We evaluated our method on two light microscopic cell datasets: MDA231 and PhC-HeLa.
MDA231 from human breast carcinoma, consists of 96 images with segmented ground truth files by experts.
The second dataset is PhC-HeLa, which consists of 22 phase contrast images of cervical cancer colonies of HeLa cells. The ground truth for this dataset consists of cell markers for all 2,228 cells.

\begin{figure}
\includegraphics[width=0.48\textwidth]{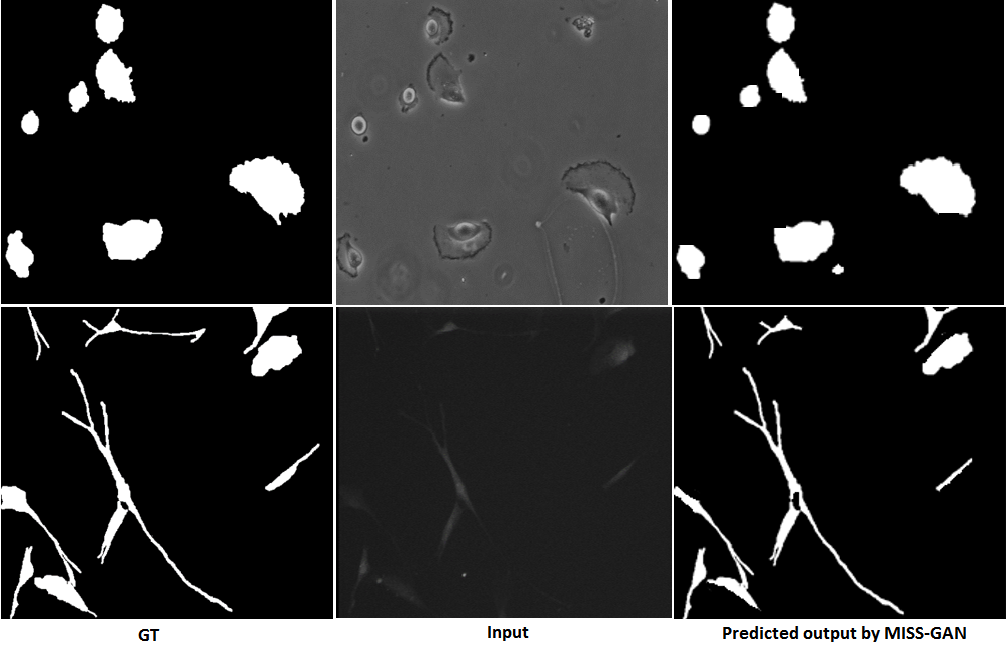}
\centering
\caption{Microscopic cell segmentation results obtained by cGAN+Refinement network with patient-wise mini-batch normalization and without Gaussian noise.}
\label{figcall}
\end{figure}

\begin{figure*}[!t]
  \centering
  \subfloat[]{\includegraphics[width=0.48\textwidth]{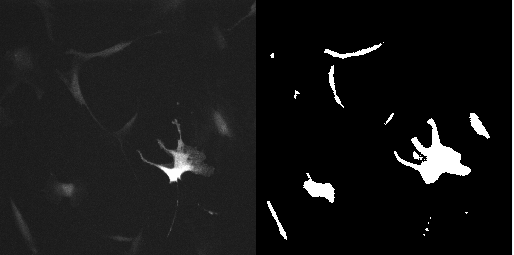}\label{figaa}}
  \hfill
  \subfloat[]{\includegraphics[width=0.48\textwidth]{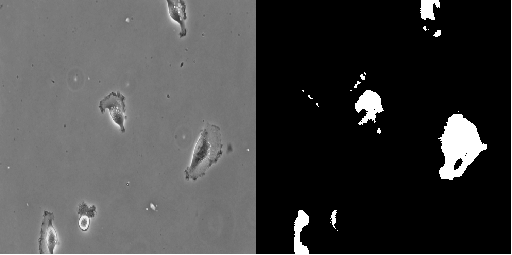}\label{figbb}}
  \caption{Microscopic cell segmentation results obtained by cGAN (a,b) without patient-wise mini-batch normalization.}
\label{fig_cellschaffel}
\end{figure*}

Figures~(\ref{figcall} and \ref{fig_cellschaffel}) compare the qualitative results from test set when the network were trained with and without patient-wise mini-batch normalization. The patient-wise mini-batch normalization provided normalization for any layer of neural network based on all available 2D images from same patient.

Based on qualitative results and Fig~(\ref{figcall}), our network is able to learn from few samples (MDA231 and PhC-HeLa) as well as large sample dataset (BraTS2017). We compared quantitative results with the state-of-the-art segmentation method. The quantitative results of individual cell segmentation are detailed in Table~(\ref{tabel_cell2}, \ref{tabel_cell1}). Obviously, we can see that diversity and the amount of images did not have a major effect on the final result.

\begin{figure*}[!t]
  \centering
  \subfloat[]{\includegraphics[width=0.48\textwidth]{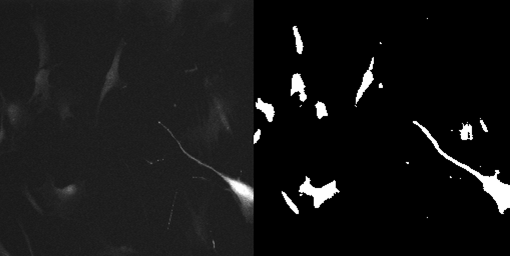}\label{figa}}
  \hfill
  \subfloat[]{\includegraphics[width=0.48\textwidth]{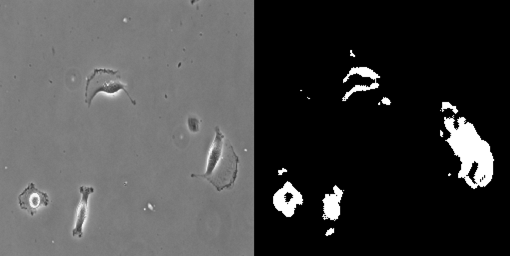}\label{figb}}
  \caption{Microscopic cell segmentation results obtained by cGAN (a,b) when the cGAN model trained  with additional Gaussian noise as input.}
\label{fig_cellnoise}
\end{figure*}

As shown in Fig.~(\ref{fig_cellnoise}) and Table (\ref{tabel_cell2}) the Gaussian noise negatively influence the segmentation results specially when the trained dataset has few samples.
We had same policy for data augmentation on all datasets. 
We explored during raining the large dataset when the generator networks takes Gaussian noise vector beside medical images, act mostly same as without noise vector and there is minimum differences in the output samples. In contrast, trained network with few samples along with noise vector has negative effect on the final outputs. 

\begin{table}[!htbp]
\centering
\caption{The achieved accuracy for cell segmentation in terms of Intersection over union on PhC-HeLa from microscopic cell data}
\begin{tabular}{*4c} 
\toprule
Approaches    &  SEG & Spec  & Sen \\
\midrule
MISS-GAN   & 0.951 & 0.943  & 0.94   \\
cGAN   &  0.928 & 0.910  & 0.91   \\
U-Net~\cite{ronneberger2015u}   &  0.92 & -  & -    \\
KTH-SE~\cite{magnusson2012batch}  &  0.79 & -  & - \\
MSER~\cite{arteta2012learning} &  0.77 & -  & - \\
Greedy~\cite{Akram}&  0.87 & -  & - \\
\bottomrule
\end{tabular}
\label{tabel_cell1}
\end{table}

\begin{table}[!htbp]
\centering
\caption{The achieved accuracy for cell segmentation in terms of intersection over union on the MDA231 data}
\begin{tabular}{*6c} 
\toprule
Approaches & SEG & Spec  & Sen & FPR & FNR  \\
\midrule
cGAN+Refinement  & 0.93 & 0.93 & 0.92 &  0.07   & 0.08   \\
RNN-GAN   &  0.91 & 0.90  & 0.91  & 0.10   & 0.09  \\
cGAN   &  0.90 & 0.89  & 0.91  & 0.11   & 0.09  \\
UNet~\cite{ronneberger2015u}   &  0.92 & -  & -  & -    \\
KTH-SE~\cite{magnusson2012batch}   &  0.79 & -  & - & -   \\
MSER~\cite{arteta2012learning} &  0.75 & -  & - & -  \\
Greedy~\cite{Akram}&  0.85 & -  & - & -  \\
\bottomrule
\end{tabular}
\label{tabel_cell2}
\end{table}
 
\section{Conclusion} \label{conclusion}
In this paper, we introduced a novel deep architecture to mitigate the issue of unbalanced data and improve the false discovery rate in medical image segmentation task.
To this end, we proposed a generator network and couple discriminator networks where a generator segments pixels label, and discriminator classifies whether segmented output is real or fake. Another discriminator called refinement network, is trained on prediction of false positive and false negative masks predicted by generator.
Moreover, we analyzed an effects of different architectural choices and a patient-wise mini-batch technique that help to improve semantic segmentation results.
Our proposed method shows outstanding results for microscopic cell segmentation and liver lesion segmentation.
We achieved competitive results in brain tumour segmentation and liver segmentation.
In the future, we plan to investigate the potential of current network for learning multiple clinical tasks such as diseases classification and semantic segmentation.

\section{Acknowledgment}
This article does not contain any studies with human participants or animals performed by any of the authors and only used the public medical dataset provided by public challenges (BraTS 2017, LiTS 2017, Microscopic cell segmentation 2015) and does not contain patient data. This article is original work that has not been published or is currently not under review at another venue. 

{\small
\bibliographystyle{ieee}
\bibliography{strings}
}

\end{document}